\documentclass[letterpaper, 10 pt, conference]{ieeeconf}  

\IEEEoverridecommandlockouts                              
\usepackage[colorlinks,linkcolor=green,citecolor=red,urlcolor=blue,bookmarks=true,hypertexnames=true]{hyperref} 
\usepackage[utf8]{inputenc}
\usepackage[T1]{fontenc}
\usepackage{float}
\usepackage{subfigure}
\usepackage{bbding}
\usepackage{balance}
\usepackage{array}
\usepackage{tabularx}

\usepackage{graphicx} 
\usepackage{epsfig} 
\usepackage{mathptmx} 
\usepackage{mathptmx} 
\usepackage{amsmath} 
\usepackage{amssymb}  
\usepackage{booktabs}
\usepackage{url}
\usepackage{multirow}
\usepackage{multicol}
\usepackage{xcolor}
\title{\LARGE \bf A Benchmark for Vision-Centric HD Mapping by V2I Systems}

\author{Miao Fan$^{1,*}$, Shanshan Yu$^{2}$, Shengtong Xu$^{3}$, Kun Jiang$^{4}$, Haoyi Xiong$^{5}$, and Xiangzeng Liu$^{6}$
\thanks{$^{1}$Chief scientist at NavInfo Co. Ltd., China. Senior member of IEEE.}
\thanks{$^{2}$Engineer at NavInfo Co. Ltd., China.}
\thanks{$^{3}$Principal product manager at Autohome Inc., China.}
\thanks{$^{4}$Associate professor at Tsinghua University, China.}
\thanks{$^{5}$Principal scientist at Baidu Inc., China. Senior member of IEEE.}
\thanks{$^{6}$Associate professor at Xidian University, China.}
\thanks{*Correspondence: {miao.fan@ieee.org}.}
}
        
\begin{document}

\maketitle
\thispagestyle{empty}
\pagestyle{empty}

\begin{abstract}
Autonomous driving faces safety challenges due to a lack of global perspective and the semantic information of vectorized high-definition (HD) maps. Information from roadside cameras can greatly expand the map perception range through vehicle-to-infrastructure (V2I) communications. However, there is still no dataset from the real world available for the study on map vectorization onboard under the scenario of vehicle-infrastructure cooperation. To prosper the research on online HD mapping for Vehicle-Infrastructure Cooperative Autonomous Driving (VICAD), we release a real-world dataset, which contains collaborative camera frames from both vehicles and roadside infrastructures, and provides human annotations of HD map elements. We also present an end-to-end neural framework (i.e., V2I-HD) leveraging vision-centric V2I systems to construct vectorized maps. To reduce computation costs and further deploy V2I-HD on autonomous vehicles, we introduce a directionally decoupled self-attention mechanism to V2I-HD. Extensive experiments show that V2I-HD has superior performance in real-time inference speed, as tested by our real-world dataset. Abundant qualitative results also demonstrate stable and robust map construction quality with low cost in complex and various driving scenes. As a benchmark, both source codes and the dataset have been released at OneDrive\footnote{\url{https://1drv.ms/f/c/76645c25a8914a0b/EgWy5XCUk6pKgvE9vB-HbVEBCdCQjJvgx1KKjeKF7hPdZw}} for the purpose of further study.
\end{abstract}
{\keywords Vehicle-to-Infrastructure (V2I), HD maps, vision-centric, benchmark.}


\section{Introduction}
High-definition (HD) maps~\cite{r0} are the most fundamental component of autonomous driving systems, providing centimeter-level details of traffic elements, vectorized topology, and navigation information. HD maps instruct the ego-vehicle to precisely locate itself on the road and anticipate what is coming up ahead. Currently, traditional SLAM-based solutions~\cite{shan2018lego, shan2020lio, zhang2014loam} have been widely adopted in practice. However, challenges such as high annotation costs and delayed updates have led to a gradual shift from offline approaches to learning-based online HD map construction using onboard sensors. Recently, online HD maps constructed in real-time around the ego vehicle using onboard sensors have effectively addressed these issues.

\begin{figure}
  \includegraphics[width=\columnwidth]{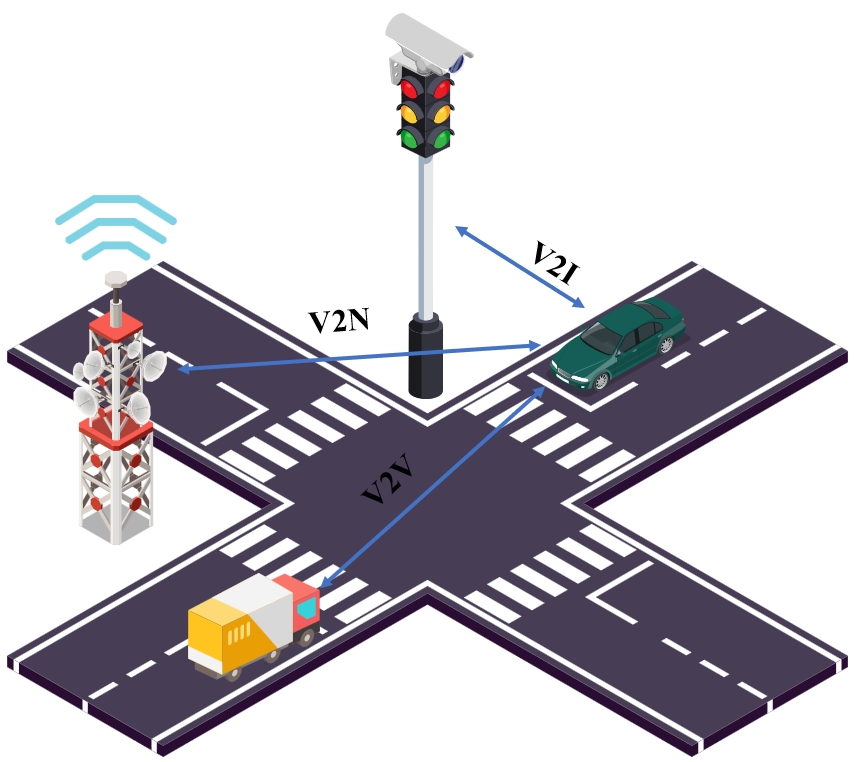}
  \caption{Cooperative systems in autonomous driving. A comprehensive autonomous driving perception system is composed of the ego vehicle and cooperative interactions among vehicles, infrastructure, and networks.}
  \label{fig:1}
    
\end{figure}

Recent works employ vehicle-mounted surround view perception and point cloud data to enable end-to-end construction of high-definition (HD) maps. Despite advances in single-vehicle perception, they are restricted to limited field of view, occlusions, and short-range perception, which result in suboptimal performance in these scenarios. Additionally, compared to offline maps, online HD map construction also encounters inherent limitations stemming from on-vehicle sensors and real-time road conditions. These challenges include variations in data quality caused by swift movement and limitations in the sensor field of view.

A promising solution to address these challenges is to leverage infrastructure information through Vehicle-to-Everything (V2X) communication, which has been proven to significantly extend perception range and improve the safety of autonomous driving, as shown in Fig.~\ref{fig:1}. Intelligent roadside infrastructure or roadside units (RSUs) equipped with sensors provide uninterrupted and continuous observations with an extensive field of view. These observations enable real-time updates to the dynamically evolving HD map, enhancing its overall perceptual accuracy. Recently, several datasets have been collected from a roadside perspective, making them particularly valuable for advancing perception algorithms in V2X systems. However, these datasets are deficient in certain map element annotations, limiting their applicability for HD map construction.

\begin{figure*}[htbp]
  \centering
  \includegraphics[width=\textwidth]{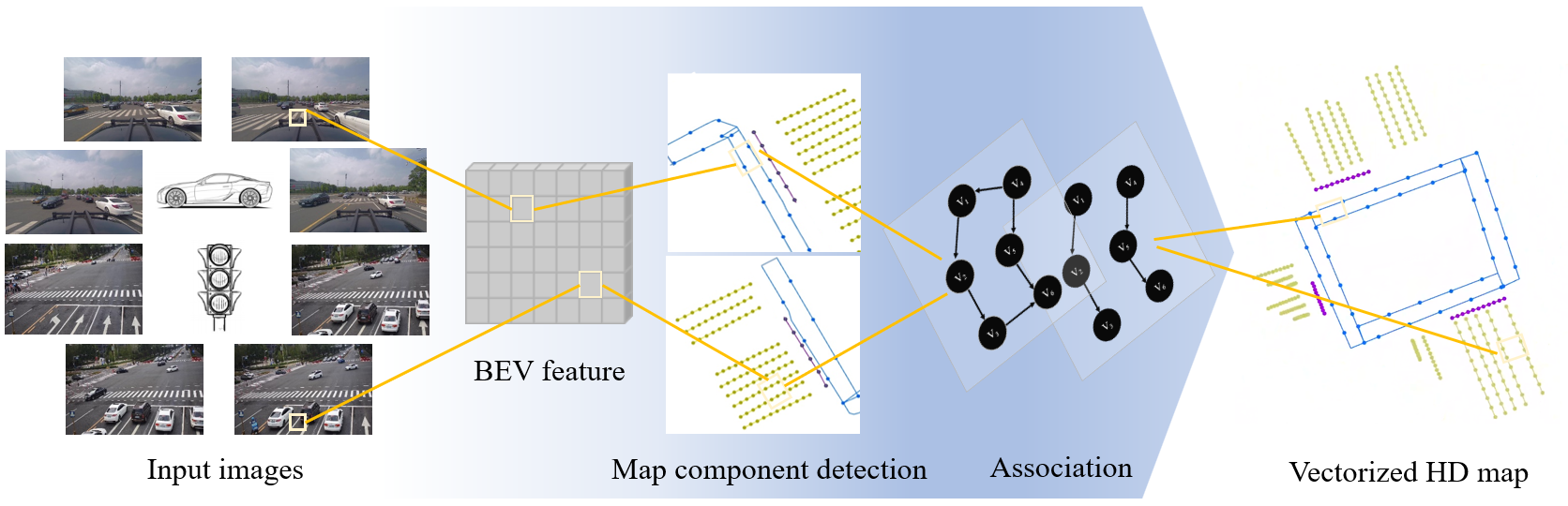}
  \caption{The pipeline of V2I-HD framework. V2I-HD is a hybrid architecture that combines CNNs with DETR, for real-time HD map learning in a BEV representation. Starting with the input of collaborative camera frames from vehicles and infrastructures, a unified BEV representation is extracted by projecting and fusing image features. The V2I-HD models the map elements through an equivalent vectorized point set, with the final vectorized HD map elements generated throughout the DETR architecture.}
  \label{fig:2}
\end{figure*}

According to the work mentioned, it can be recognized that unified map element annotations of camera images from the roadside perspective encompassing diverse traffic participants and scenarios remain rare. The DAIR-V2X-Seq dataset~\cite{yu2023v2x} encompassing 2D/3D object annotations sampled from the real world, designed for trajectory tracking and prediction in vehicle-infrastructure collaboration. However, with the lack of vectorized annotations for HD map elements, we release a dataset inherited from DAIR-V2X-Seq, which encompasses corresponding HD map element annotations by cropping vector maps from frames captured by vehicle-mounted and roadside cameras, making it appropriate for HD map construction tasks in V2I contexts. To our knowledge, the transformed dataset is the first release that focuses on HD map construction, making it an ideal resource for the development and evaluation of cooperative perception. 

We further propose a novel method called V2I-HD, which copes with creating HD maps onboard for autonomous driving at a minimal cost while achieving superior state-of-the-art performance. In our design, infrastructure cameras from a bird's eye view cooperate with the front-view image of a vehicle~\cite{r00} to construct HD maps. V2I-HD initially extracts features from both vehicle\&infrastructure-side images, and subsequently transforms these features into a cohesive BEV representation via a map encoder. Then, we leverage the map decoder to estimate and update the map topology. The map decoder comprises map queries and decoder layers. Each decoder layer updates the map query utilizing a direction-decoupled attention scheme.

The contributions of this paper are thus the following:
\begin{itemize}
    \item We release a real-world dataset, which contains collaborative camera frames from both vehicles and roadside infrastructures and provides HD map element annotations.

    \item We propose a structured end-to-end framework for efficient online vectorized HD map construction, building on DETR. To reduce computation costs, we introduce directionally decoupled self-attention.

    \item We demonstrate the capabilities of the solution on the map and traffic data, and conduct a quantitative assessment of the algorithm to a sub-optimal methods.
\end{itemize}

This work is structured as follows: Section II discusses related work. Section III presents the problem formulation, neural architecture, and our training strategy. Implementation and experiments are described in sections IV. Finally, section V gives a conclusion.

\section{Related Work}

This section reviews related work concerning HD map construction under Vehicle-to-Infrastructure (V2I) communication, deriving information from available datasets.

\subsection{Vectorized HD Map Construction}
HD maps consist of geometric objects and semantic properties, both of which are crucial for downstream tasks. HD map construction in BEV space~\cite{ding2023pivotnet, hao2024mbfusion} relies on data gathered by onboard sensor observations, encompassing RGB images from multi-view cameras and point clouds from LiDAR. Current methodologies for HD map construction can be broadly classified into two categories: rasterized HD map estimation~\cite{li2022hdmapnet, li2022bevformer, liu2023bevfusion, xiong2023neural} and vectorized HD map construction~\cite{ding2023pivotnet, liao2022maptr, liu2023vectormapnet, qiao2023end}. Rasterized HD maps often require extensive post-processing, making them less ideal for downstream tasks. In contrast, vectorized HD map construction addresses these constraints by representing maps using a collection of map elements. For instance, VectorMapNet~\cite{liu2023vectormapnet} explores the keypoint-based representation within a hierarchical two-stage network. InsightMapper~\cite{xu2023insightmapper} demonstrates the benefits of leveraging internal instance point data. The MapTR series~\cite{liao2022maptr, liao2023maptrv2} introduces permutation-equivalent modeling of point sets and a DETR-like~\cite{carion2020end} single-stage network. Recent works have focused on learning element-level information. For example, MapVR~\cite{zhang2024online} incorporates differentiable rasterization and enhances supervision through element-level segmentation. BeMapNet~\cite{qiao2023end} initially identifies map elements and subsequently enhances detailed nodes with a segmented Bezier head. PivotNet~\cite{ding2023pivotnet} proposes a Point-to-Line Mask module, which transforms point-level representations into element-level representations. 

\begin{figure}[bp]
\centering
\subfigure[Red Light Violation]{\includegraphics[height=3cm,width=4.2cm]{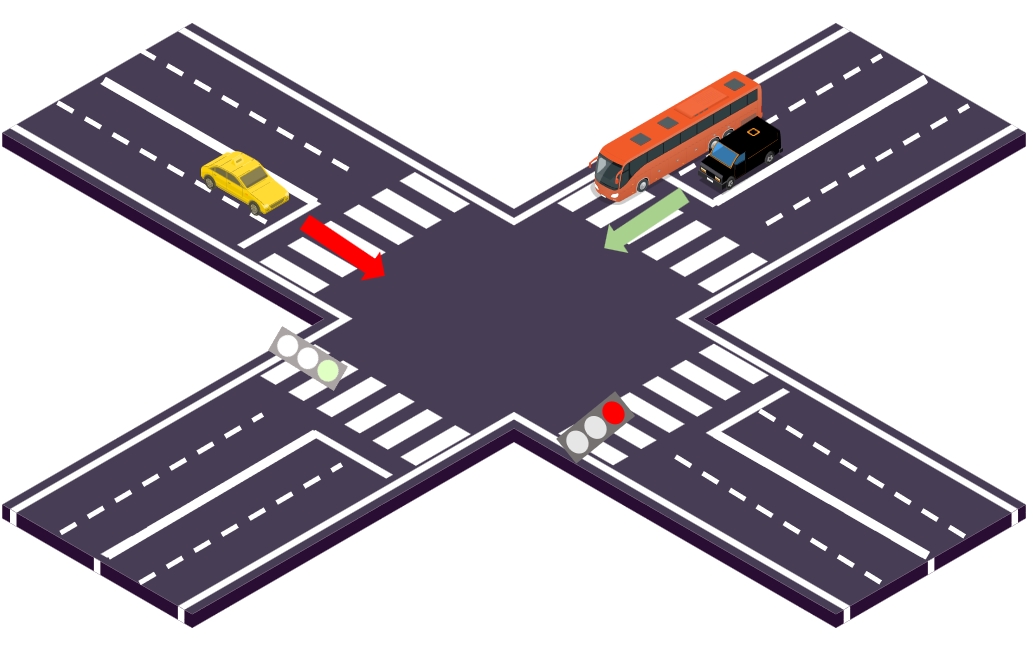}}
\subfigure[Unprotected Left Turn]{\includegraphics[height=3cm,width=4.2cm]{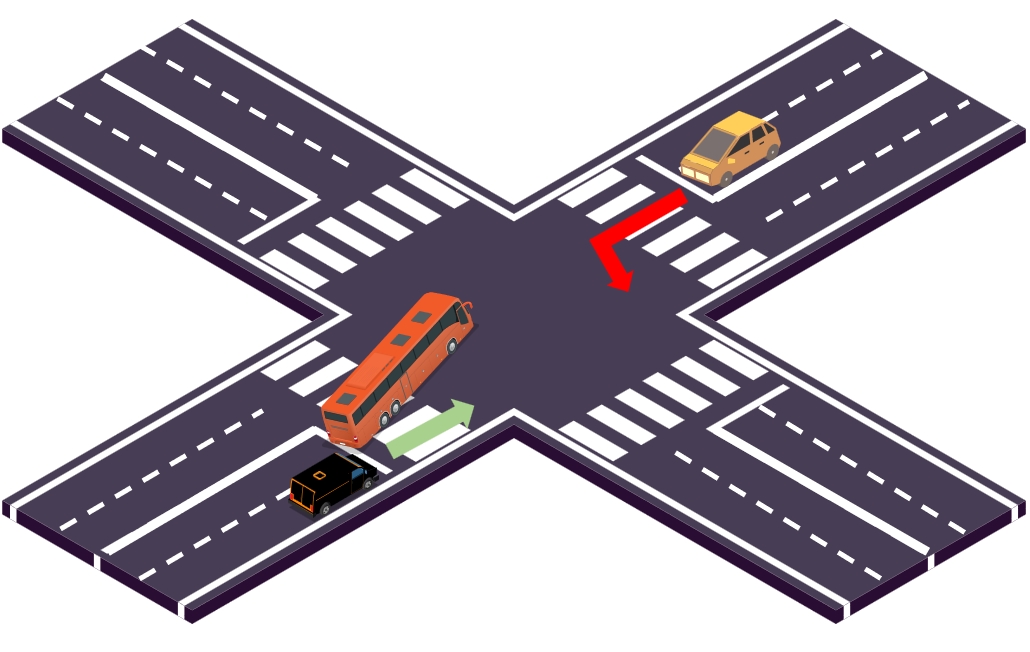}}
\caption{The red truck(s) block the black vehicle’s view. It cannot see the oncoming yellow vehicle violating the red light in (a) and taking the unprotected left turn in (b).}
\label{fig:3}
\end{figure}

\begin{figure*}[t]
  \centering
  \includegraphics[width=\textwidth]{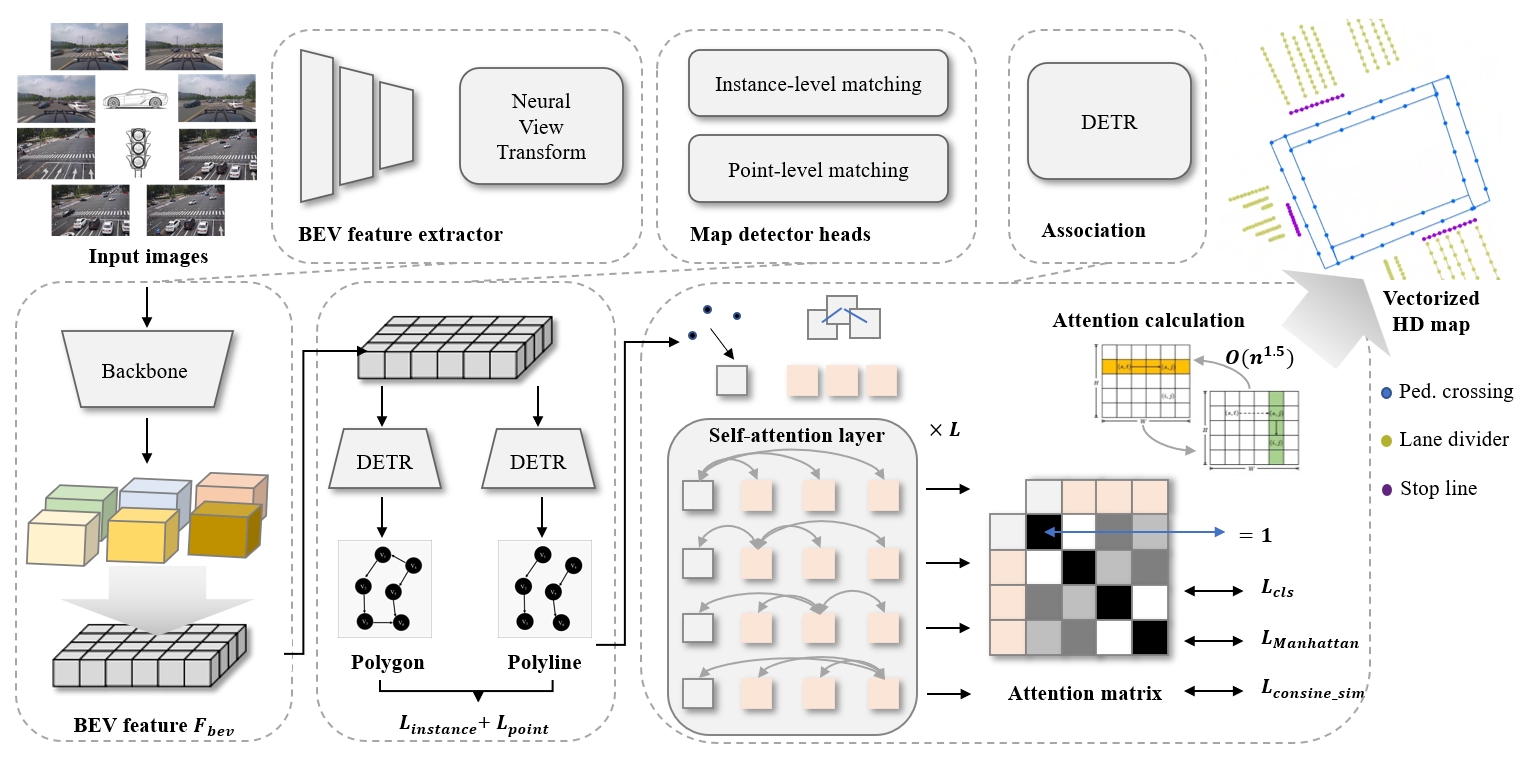}
  \caption{The framework of V2I-HD . The upper blocks show the main components of V2I-HD, and the lower blocks provide detailed information regarding the structure and training of each component.}
  \label{fig:4}

\end{figure*}

\begin{table*}[htp]
\caption{Datasets used for online mapping. }
\begin{center}
\setlength{\tabcolsep}{4.6mm}
\begin{tabularx}{\linewidth}{l>{\centering\arraybackslash}X>{\centering\arraybackslash}X>{\centering\arraybackslash}p{2.6cm}>{\centering\arraybackslash}X>{\centering\arraybackslash}X>{\centering\arraybackslash}X>{\centering\arraybackslash}X}
\toprule
\multirow{2}{*}{\textbf{Dataset}} & 
\multirow{2}{*}{\textbf{Split}} & 
\multirow{2}{*}{\textbf{Source}} &
\multirow{2}{*}{\textbf{Main Map Purpose}} &
\multicolumn{3}{c}{\textbf{\#Samples}}&
\multirow{2}{*}{\textbf{Geo.Split}} \\
\cmidrule(lr){5-7}
\multicolumn{2}{c}{} & & &Train & Valid &Test & \\
\midrule
nuScenes & Original & nuSc & OD/MF & 28K & 6K & 6K & \XSolidBrush \\
Argoverse 1 & Original & argo1 & OD/MF & 39K & 15K & 13K & \Checkmark \\
Argoverse 2 & Original & argo2 & OD/MF & 110K & 24K & 24K & \XSolidBrush \\
Waymo & Original & way & OD/MF & 122K & 30K & 40K & \XSolidBrush \\
\midrule
nuScenes & Near & nuSc & OM & 28K & 6K & 6K & \Checkmark \\
Argoverse 2 & Near & argo2 & OM & 110K & 24K & 24K & \Checkmark \\
\midrule
nuScenes & Far-A & nuSc & OM & 30K & 9K & - & \Checkmark \\
nuScenes & Far-B & nuSc & OM & 31K & 9K & - & \Checkmark \\
Argoverse 2 & Far-A & argo2 & OM & 110K & 46K & - & \Checkmark \\
Argoverse 2 & Far-B & argo2 & OM & 101K & 55K & - & \Checkmark \\
Argoverse 2 & Far-C & argo2 & OM & 101K & 55K & - & \Checkmark \\
\bottomrule
\vspace{0.1\baselineskip} \\
\end{tabularx}
\label{tab:1}
\parbox{\linewidth}{Datasets used for HD map construction. The proposed splits are shown in bold. OD = object detection, MF = motion forecasting, OM = online mapping.}
\end{center}
\end{table*}

However, all of the above work relies on HD map constructed by sensors from the vehicle ego, whose perception has long suffered from limitations of range restrictions, sensor field of view, and occlusions. The National Highway Transportation and Safety Authority (NHTSA) has outlined several scenarios in which occlusions may result in traffic collisions \cite{najm2013description}. In one such scenario (Fig. \ref{fig:3}), the black vehicle, equipped with a front view camera, will collide with an oncoming yellow vehicle that is disregarding a red light, as a red truck obstructs its frontal view. If the black vehicle had seen the oncoming vehicle, it might have averted the collision. A potential solution to this problem is the implementation of intelligent roadside infrastructure or Roadside Units (RSUs) to enhance the vehicle's sensor range. Early work from He \cite{he2023vi} utilized roadside infrastructure to enhance the vehicle's field of vision, facilitating real-time map inference. Compared with other state-of-the-art methods of online HD inferencing, this solution significantly improved the safety of autonomous driving. In this study, we propose a structured end-to-end framework for the development of HD maps to enhance both accuracy and coverage. Different from He's method, our method employs only roadside cameras and the vehicle’s front-facing camera.

\subsection{Available Datasets}
Table \ref{tab:1} presents a summary of the datasets utilized for the construction of HD maps. Four original datasets, specifically Argoverse 1 and 2 \cite{chang2019argoverse, wilson2023argoverse}, nuScenes \cite{caesar2020nuscenes}, Waymo \cite{sun2020scalability}, and DAIR-V2X-Seq \cite{yu2023v2x}, provide the HD map data necessary for online map inference. These datasets with comprehensive object annotations are mainly designed for object tracking and trajectory prediction tasks, encompassing vehicle-point clouds and images. Although these publicly available datasets facilitate fair and consistent evaluation in research, they were initially designed for dynamic object perception rather than HD map inference. To avert sample overlap within a sequence, the datasets are temporally divided but do not ensure geographic separation. Despite this, nuScenes \cite{caesar2020nuscenes} and Argoverse 2 \cite{wilson2023argoverse} are widely used for training online mapping models and have become the de facto standard. Nevertheless, numerous research have employed nuScenes or Argoverse 2 to train online mapping models, which have emerged as de facto standards in the domain. For example, online mapping methods using nuScenes include \cite{hu2022st, jiang2023polarformer, li2022bevformer, li2022hdmapnet} and Argoverse 2 is used in \cite{liao2022maptr, liao2023maptrv2}. Both nuScenes and Argoverse 2 are devoid of data from roadside sensors, hence constraining their use in V2I scenarios. DAIR-V2X-Seq encompasses infrastructure-based data; nevertheless, this dataset is tailored for object tracking and trajectory prediction tasks, rendering it unapplicable for map inference. This work releases a dataset specifically created for the online inference of HD maps, featuring annotations of map components related to both vehicle-side and roadside images.

\section{Methods}

Here, we describe how we model the problem, construct our network, and further minimize computational cost.

\subsection{Problem Formulation}

By sampling the sequence of components in the vector map, each feature can be represented as curves of varying shapes. Following the methodology presented in MapTRv2 ~\cite{liao2023maptrv2}, we encapsulate map features into closed geometries (e.g., crosswalks) and open geometries (e.g., divider lines, stop lines). Uniform sequential sampling along feature boundaries abstracts the geometric representation of closed shapes as polygons and open shapes as polylines.

Each map element corresponds to $\mathcal{V}=(V, \Gamma)$.$V=\left\{v_j\right\}_{j=0}^{N_v-1}$ denotes a collection of points of the map element($N_v$ is the number of points). $\Gamma = {\gamma ^k}$ signifies a group of equivalent permutations of the point set $V$, including all potential permutations of the specified map feature. Specifically, for polyline element with unspecific direction, $\Gamma$ includes 2 kinds of equivalent permutations:

\begin{equation}
\begin{aligned}
\Gamma_{\text {polyline }} & =\left\{\gamma^0, \gamma^1\right\} \\
& =\left\{\begin{array}{l}
\gamma^0(j)=j \% N_v \\
\gamma^1(j)=\left(N_v-1\right)-j \% N_v
\end{array}\right.
\end{aligned}
\end{equation}

For a polyline element with a specified direction, $\Gamma$ includes only one permutation: ${\gamma^0}$.

For polygon element, $\Gamma$ includes $2×N_v$ kinds of equivalent permutations:
\begin{equation}
\begin{aligned}
\Gamma_{\text {polygon }}= & \left\{\gamma^0, \ldots, \gamma^{2 \times N_v-1}\right\} \\
= & \left\{\begin{array}{l}
\gamma^0(j)=j \% N_v \\
\gamma^1(j)=\left(N_v-1\right)-j \% N_v \\
\gamma^2(j)=(j+1) \% N_v \\
\gamma^3(j)=\left(N_v-1\right)-(j+1) \% N_v \\
\ldots \\
\gamma^{2 \times N_v-2}(j)=\left(j+N_v-1\right) \% N_v \\
\gamma^{2 \times N_v-1}(j)=\left(N_v-1\right)-\left(j+N_v-1\right) \% N_v
\end{array}\right.
\end{aligned}
\end{equation}

\subsection{Neural Architecture}
The comprehensive model architecture is depicted in detail in Fig. \ref{fig:2}, which delineates the framework into three components: feature extractor, Map Encoder, and Map Decoder.

\subsubsection{Feature extractor} With inputs of images from the vehicle's front-facing camera and the overhead roadside view, we initially extract features from each image employing a common CNN backbone. Then, multi-scale features from various phases are fed into a Feature Pyramid Network (FPN) \cite{tan2020efficientdet} to integrate comprehensive semantic information. Ultimately, we upsample the pyramid features to a uniform size and stack them as the final output.

\subsubsection{Map encoder} We utilize a conventional Transformer-based architecture to consistently convert image attributes into the BEV space. The BEV decoder models the task as a set prediction problem using perspective transformation, which takes camera features with shape $H_c \times W_c$ and $H_q \times W_q$ queries as inputs and produces BEV features $F_b \in \mathbb{R}^{C \times H_b \times W_b}$  by modeling all pairwise interactions among elements with self-attention. Currently, other PV2BEV approaches exist, e.g., CVT \cite{zhou2022cross}, LSS \cite{philion2020lift}, Deformable Attention \cite{li2022bevformer}, GKT \cite{chen2022efficient} and IPM \cite{mallot1991inverse}; however, given that our model must be implemented on the vehicle side, we have chosen GKT as the default transformation method.

\subsubsection{Map decoder} 
\begin{figure}
  \includegraphics[width=\columnwidth]{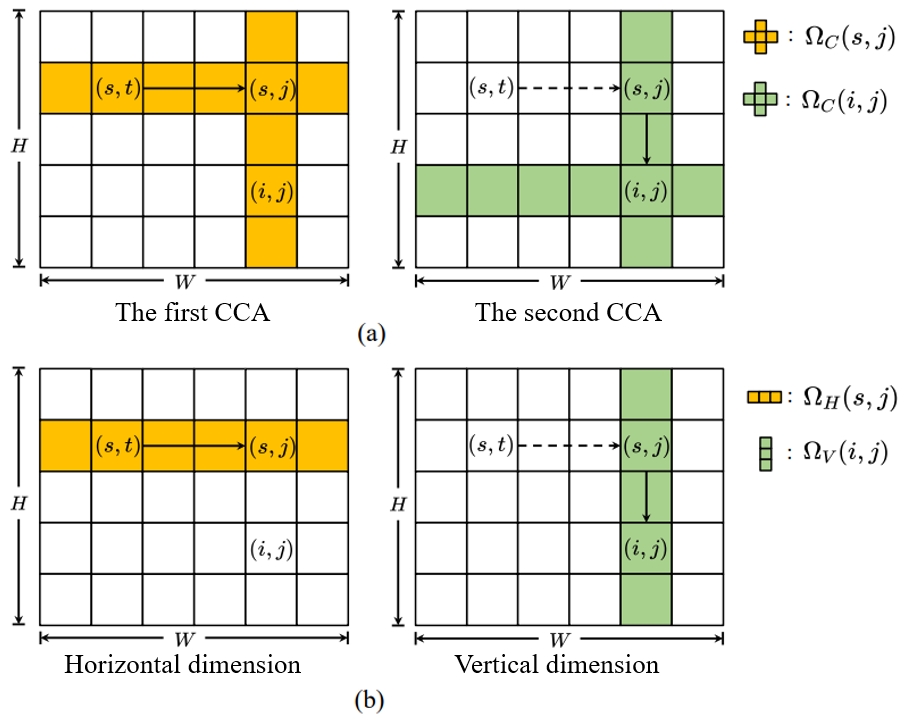}
  
  \caption{The difference between the attention component of RCCA and a non-local similarity algorithm using direction decoupling.}
  \label{fig:5}

\end{figure}

The map decoder comprises map queries and many decoder layers, with each layer enhancing the element queries via attention mechanisms. Current methodologies modify queries utilizing raw attention, leading to a computational complexity of $O((N \times N_v)^2$, where $N$ and $N_v$  denote the quantities of instance queries and point queries, respectively. As the volume of requests escalates, the computing expense rises rapidly. To alleviate this computational burden, we employ direction-decoupled self-attention, which initially calculates attention in the horizontal direction followed by the vertical direction. This method enables the vertical attention computation to integrate information from the horizontal direction, as depicted in Figure \ref{fig:5}. Decoupled self-attention decreases the computational complexity from $O((N \times N_v)^2$ to $O((N \times N_v)^{1.5}$ ~\cite{9930662} and outperforms the conventional self-attention technique.

\subsubsection{Output head} Utilizing the methodology described in the problem modeling section, we construct a segmented output head that incorporates both instance matching and point-level matching. Initially, the instance class scores are forecasted, succeeded by the regression of point-level distance losses. The outputs are produced by amalgamating these predictions, resulting in a vector of dimension $2 N_{v^{-}}$ or $2 N_{v^{-}}$, which signifies normalized 2D or 3D coordinates of the $N_v$ points. 

\subsection{End-to-End Training}
\begin{figure*}[t]
  \centering
  \includegraphics[width=\textwidth]{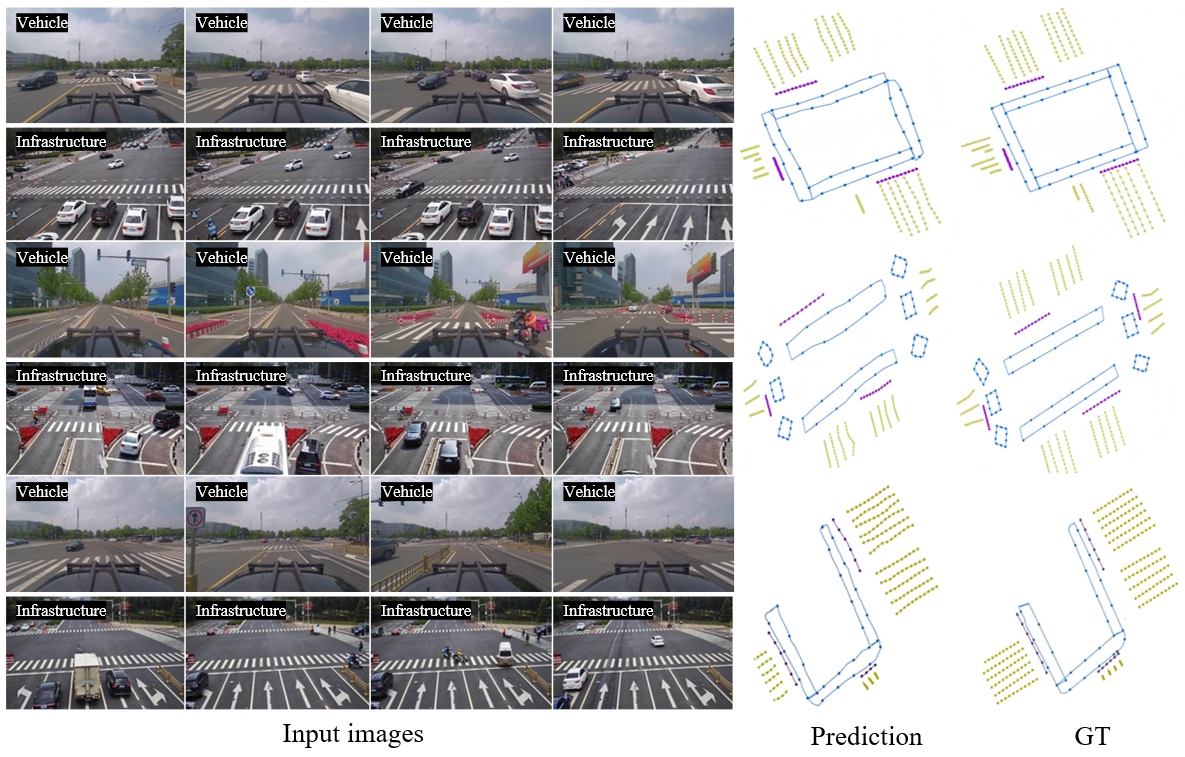}
  
  \caption{Qualitative results of V2I-HD on complex traffic scenes under various conditions. Lane dividers, stoplines, and pedestrian crossings are visualized in yellow, blue, and purple.}
  \label{fig:6}

\end{figure*}

\subsubsection{Ground truth} The DAIR-V2X-Seq dataset is deficient in camera depth parameters, rendering direct projection of the pictures into the world coordinate system unfeasible. To address this problem, we correlate the point cloud data with the image data to generate depth maps, which are subsequently employed to align the images with the HD map in the world coordinate system, facilitating the annotation of map features. The map feature annotations are efficiently represented as curves utilizing a collection of vector points. These annotations data are released via our dataset.

\subsubsection{Loss function}
The framework is trained via instance matching and point set regression. The fundamental loss function comprises three components: classification loss, point-to-point loss, and edge direction loss:
\begin{equation}
\begin{aligned}
\mathcal{L}_{\text {one2one }} & =\mathcal{L}_{\text {Hungarian }}(\hat{Y}, Y) \\
& =\lambda_c \mathcal{L}_{\mathrm{cls}}+\lambda_p \mathcal{L}_{\mathrm{p} 2 \mathrm{p}}+\lambda_d \mathcal{L}_{\mathrm{dir}}
\end{aligned}
\end{equation}

Each predicted map element is designated a class label based on the instance-level optimal matching outcome. The classification loss is defined as a Focal Lossterm formulated as:
\begin{equation}
\mathcal{L}_{\mathrm{cls}}=\sum_{i=0}^{N-1} \mathcal{L}_{\text {Focal }}\left(\hat{p}_{\hat{\pi}(i)}, c_i\right) .
\end{equation}

Point-to-point loss regulates the position of each predicted point. For each ground truth (GT) instance indexed by $i$, based on the point-level optimal matching result $\hat{\gamma_i}$, each predicted point $\hat{v}_{\hat{\pi}(i), j}$ is allocated to a GT point $v_{i, \hat{\gamma}_i(j)}$. The point-to-point loss is defined as the Manhattan distance calculated between each assigned point pair:
\begin{equation}
\mathcal{L}_{\mathrm{p} 2 \mathrm{p}}=\sum_{i=0}^{N-1} \mathbb{1}_{\left\{c_i \neq \varnothing\right\}} \sum_{j=0}^{N_v-1} D_{\text {Manhattan }}\left(\hat{v}_{\hat{\pi}(i), j}, v_{i, \hat{\gamma}_i(j)}\right) .
\end{equation}

Point-to-point loss exclusively supervises the vertex of the polyline and polygon, disregarding the edge. The orientation of the edge is crucial for the precise representation of map elements. Consequently, we furthermore formulate edge direction loss to regulate the geometric configuration at the elevated edge level:
\begin{equation}
\mathcal{L}_{\mathrm{dir}}=-\sum_{i=0}^{N-1} \mathbb{1}_{\left\{c_i \neq \varnothing\right\}} \sum_{j=0}^{N_v-1} \operatorname{cos\_ similarity}\left(\hat{\boldsymbol{e}}_{\hat{\boldsymbol{\pi}}(i), j}, \boldsymbol{e}_{\boldsymbol{i}, \hat{\gamma}_i(j)}\right),
\end{equation}

\begin{table*}[htp!]
\caption{Evaluation results on V2X seq dataset.}
\setlength{\tabcolsep}{4.6mm}
\begin{center}
\begin{tabularx}{\linewidth}{l>{\centering\arraybackslash}X>{\centering\arraybackslash}p{1.3cm}>{\centering\arraybackslash}X>{\centering\arraybackslash}p{1.0cm}>{\centering\arraybackslash}X>{\centering\arraybackslash}X>{\centering\arraybackslash}X>{\centering\arraybackslash}X}
\toprule
\multirow{2}{*}{\textbf{Method}} & 
\multirow{2}{*}{\textbf{Modality}} & 
\multirow{2}{*}{\textbf{Backbone}} &
\multirow{2}{*}{\textbf{Epochs}} &
\multicolumn{3}{c}{\textbf{AP}} & 
\multirow{2}{*}{\textbf{mAP}} &
\multirow{2}{*}{\textbf{FPS}} \\
\cmidrule(lr){5-7}
& & & & pedestrian crossing & lane divider & stop line & & \\
\midrule
HDMapNet~\cite{li2022hdmapnet} & V & EffiNet-B0 & 30 & 12.0 & 5.2 & 18.9 & 12.0 & - \\
HDMapNet~\cite{li2022hdmapnet}  & V\&I & EffiNet-B0 & 30 &  14.8 & 8.1 & 23.3 & 15.5 & - \\
\midrule
MapTR~\cite{liao2022maptr} & V & R18 & 110 & 21.6 & 23.5 & 23.8 & 22.9 & 38.2\\
MapTR~\cite{liao2022maptr}   & V\&I & R18 & 110 & 32.8 & 38.4 & 39.5 & 36.9 & 10.5 \\
MapTR~\cite{liao2022maptr}  & V & R50 & 110 & 26.8 & 32.1 & 34.6 & 31.1 & 16.5\\
MapTR~\cite{liao2022maptr}   & V\&I & R50 & 110 & 36.7 & 42.2 & 43.5 & 40.8 & 6.4 \\
\midrule
V2I-HD & V & R18 & 110 & 24.5 & 26.0 & 27.2 & 25.9 & 44.5\\
V2I-HD  & V\&I & R18 & 110 & 36.8 & 39.6 & 40.2 & 38.9 & 18.8 \\
V2I-HD & V & R50 & 110 & 28.4 & 30.5 & 32.8 & 30.5 & 20.8\\
V2I-HD  & V\&I & R50 & 110 & 41.4 & 45.5 & 47.8 & 44.9 & 9.6 \\
\bottomrule
\vspace{0.1\baselineskip} \\
\end{tabularx}
\label{tab:2}
\parbox{\linewidth}{Performance comparison with baseline methods at V2I sence on V2X Seq provided by the HD map construction challenge. 'V' denotes input data originating solely from the vehicle, while 'V\&I' refers to input data that includes contributions from both the vehicle and the infrastructure. The quantitative findings demonstrate that our V2I-HD substantially enhances HD map production and surpasses baseline methods.}
\end{center}
\end{table*}

\begin{table*}[htp!]
\caption{Ablation on Attention Calculation.}
\setlength{\tabcolsep}{4.6mm}
\begin{center}
\begin{tabularx}{\linewidth}{l>{\centering\arraybackslash}X>{\centering\arraybackslash}X>{\centering\arraybackslash}X>{\centering\arraybackslash}X>{\centering\arraybackslash}X>{\centering\arraybackslash}X>{\centering\arraybackslash}X}
\toprule
\multirow{2}{*}{\textbf{Self-Attn}} & 
\multirow{2}{*}{\textbf{Backbone}} & 
\multicolumn{3}{c}{\textbf{AP}} &
\multirow{2}{*}{\textbf{mAP}} &
\multirow{2}{*}{\textbf{GPU memory}} & 
\multirow{2}{*}{\textbf{FPS}} \\
\cmidrule(lr){3-5}
& & pedestrian crossing & lane divider & stop line & & & \\
\midrule
Vanilla & R18 & 26.1 & 28.4 & 30.0 & 28.1 & 11621 MB & 15.3 \\
Vanilla & R50 & 28.8 & 30.9 & 32.3 & 30.6 & 13952 MB & 15.3 \\
\midrule
RCCA & R18 & 32.7 & 34.5 & 36.0 & 34.4 & 9753 MB & 15.5 \\
RCCA & R50 & 36.9 & 39.5 & 40.5 & 38.9 & 12790 MB & 15.5 \\
\midrule
Decoupled & R18 & 36.8 & 39.6 & 40.2 & 38.9 & 8930 MB & 18.8 \\
Decoupled & R50 & 40.4 & 44.5 & 46.8 & 43.9 & 10682 MB & 9.6 \\
\bottomrule
\vspace{0.1\baselineskip} \\
\end{tabularx}
\parbox{\linewidth}{Ablation of the self-attention variations. The inter self-attention significantly decreases memory consumption while maintaining comparable accuracy. Consequently, we have designated the disconnected self-attention as the standard configuration.}
\end{center}
\label{tab:3}
\end{table*}

\begin{table}[htp!]
\caption{Ablation on BEV extractor.}
\begin{center}
\begin{tabularx}{\linewidth}{l>{\centering\arraybackslash}X>{\centering\arraybackslash}X>{\centering\arraybackslash}X>{\centering\arraybackslash}X}
\toprule
\textbf{Backbone} &
\textbf{Method} & 
\textbf{mAP} & 
\textbf{FPS}& 
\textbf{Param} \\
\midrule
\multirow{3}{*}{R18}
& IPM & 17.2 & 15.4 & 35.7 \\
& LSS & 19.6 & 12.8 & 39.2 \\
& GKT & 20.8 & 15.2 & 36.8 \\
\midrule
\multirow{3}{*}{R50}
& IPM & 17.2 & 15.4 & 35.7 \\
& LSS & 19.6 & 12.8 & 39.2 \\
& GKT & 20.8 & 15.2 & 36.8 \\
\bottomrule
\\
\end{tabularx}
\parbox{\linewidth}{Ablation on the BEV extractor. We use the classic models in the BEV feature extractor, such as LSS, and GKT. Considering that the model needs to be deployed on the vehicle side in the future, GKT is used as the conversion module. }
\end{center}
\label{tab:4}
\end{table}

\section{Experiments}
\subsection{Experimental Settings}
\subsubsection{Dataset} To evaluate the proposed approach, our dataset comprises 10669/4585 samples for the training/validation set from 28 intersections. Each scene consists of around 545 samples, with each sample including 8 images, including 4 frames from the vehicle's front-facing camera and 4 frames from the roadside overhead camera. To provide an equitable assessment, we concentrate on three static map classifications, as delineated in other studies: lane dividers, stop lines and pedestrian crossings. The perceptual range centered on the ego vehicle is established at [30, 30, 15, 15] meters, representing the distances to the front, rear, left, and right, respectively. Furthermore, we establish the resolution of the ego-to-pixel transformation at 0.15 m/pixel.

\subsubsection{Evaluation metrics} We evaluate the quality of map production by Average Precision (AP). Average Precision (AP) is determined by assessing the chamfer distance between a ground truth and a forecasted value, with a prediction deemed a true positive only if its distance falls below a specified threshold. In our trials, this threshold is established at [0.2, 0.5, 1.0] meters. The mean Average Precision (AP) is calculated by averaging the AP across the three mapping categories.

\subsubsection{Implementation details} We utilize ResNet-50/ResNet-18 and DETR as backbones, both initialized using ImageNet pretraining. The semantic BEV decoder comprises two transformer encoder layers, with default quantities of 50 for queries, 20 for point queries, and 6 for decoder layers. The input image dimensions are adjusted to $1920 \times 1080$, with a mini-batch size of 1 per RTX 3090 GPU. We train our model using 1 RTX 3090 GPU for $30/110$ epochs and implement a multi-step schedule with milestones at $[0.7, 0.9]$ and $\gamma=\frac{1}{3}$. The Adam optimizer utilizes a weight decay of $1 \times 10^{-4}$ and a learning rate of $2 \times 10^{-4}$, which is subsequently multiplied by $0.1$ for the backbone. We assigned the hyper-parameters for loss weight as follows: $\lambda_s = 1$, $\lambda_z = 5$, $\lambda_p = 5$, $\lambda_c = 10$, and $\lambda_r = 1$. Additionally, the dilated width $\omega$ in $\mathcal{L}_{\text {region }}$ was set to 5.

\subsection{Comparison Results} As our research is the inaugural exploration of using both automobiles and infrastructure for the creation of high-definition (HD) maps, there are presently no recognized competing methodologies. We additionally adapt existing vehicle-based models to the V2I (vehicle-to-infrastructure) context for vectorized HD map production to facilitate comparison with our approach. Table \ref{tab:2} presents the use of HDMapNet \cite{li2022hdmapnet}, a leading model for the generation of vectorized HD maps. Our V2I-HD conducts uncertainty-aware fusion on the static BEV (Bird's Eye View) features derived from the open-source HDMapNet model. The findings indicate that V2I-HD enhances HD map generation quality by more than 5 mAP, illustrating its efficacy in producing superior HD maps within the V2I context. 
Furthermore, we evaluate V2I-HD in terms of frames per second (FPS), which also utilizes the HD map production inference speed presented in Table \ref{tab:2}. Our V2I-HD pipeline demonstrates superior performance in both absolute improvement (43.9 mAP compared to 42.8 mAP and 15.5 mAP) and relative improvement (13.9 mAP compared to 11.7 mAP and 3.45 mAP).

We show the vectorized HD map predictions in Fig. \ref{fig:6}. It illustrates that V2I-HD has strong generalization across diverse situations. V2I-HD calculates, in an end-to-end method without post-processing or intensive computation, the semantic and instance-level data of intricate map elements. Our vector map modeling technique, illustrated in Fig. \ref{fig:4}, facilitates precise and rapid prediction of map elements.

\subsection{Ablation Study}
This section presents ablation experiments to evaluate the efficacy of the proposed modules and design decisions. All trials are run on our dataset to guarantee a fair comparison, with training spanning 110 epochs. ResNet18 and ResNet50 serve as the picture backbones in the tests.

\subsubsection{Ablation on the self-attention variants} Table \ref{tab:3} delineates the efficacy of several computational methodologies in the semantic decoder for map generation. The findings utilizing ResNet50 indicate that RCCA ~\cite{huang2019ccnet} markedly decreases training memory use (by 1162M) with just a minimal reduction in accuracy (a decline of 0.2 mAP). Moreover, in comparison to vanilla self-attention ~\cite{zhou2018atrank}, decoupled self-attention exhibits greater memory efficiency (decreasing memory use by 2108M), enhances accuracy (increasing by 5 mAP), and preserves comparable speed. Within the same computational parameters, detached self-attention attains inference with markedly less memory usage and accelerated processing speed while maintaining predictive accuracy.

\subsubsection{Ablation on BEV extractor} Table \ref{tab:4} presents the approaches for transforming 2D to BEV, encompassing IPM, LSS, and GKT, as well as the application of deformable attention. We employed the optimized implementation of LSS for the trials. To facilitate an even comparison with IPM and LSS, both GKT and deformable attention were configured using decoupled parameters. The findings indicate that V2I-HD is compatible with multiple 2D to BEV methodologies and maintains consistent performance across all configurations.

\section{Conclusion}
In this paper, we release the first dataset for online vectorized HD map construction by vision-centric V2I systems. It contains annotations for both vehicle-side and infrastructure map elements, where all the data elements are captured from the real world. To provide a benchmark of V2I HD mapping, we present a structured end-to-end framework (i.e., V2I-HD) for efficient vectorized HD map construction onboard leveraging collaborative camera frames from both vehicles and roadside infrastructures. We also introduce 
a directionally decoupled self-attention mechanism to V2I-HD for the sake of reducing computation costs. Extensive experiments show that V2I-HD has superior performance in map elements of arbitrary shape compared to other methods on our dataset.




\section*{Acknowledgments}
This work was sponsored by Beijing Nova Program (No. 20240484616). 
\balance
\bibliographystyle{IEEEtran}
\bibliography{IEEEexample}
\end{document}